\documentclass[conference]{IEEEtran}
\IEEEoverridecommandlockouts
\ifCLASSOPTIONcompsoc
  \usepackage[nocompress]{cite}
\else
  \usepackage{cite}
\fi
\ifCLASSINFOpdf
\else
\fi

\usepackage{amsmath,amssymb,amsfonts}
\usepackage{algorithmic}
\usepackage{graphicx}
\usepackage{textcomp}
\usepackage{xcolor}
\usepackage{booktabs}
\usepackage{enumitem}
\usepackage[misc]{ifsym}
\usepackage{multirow}
\usepackage{subfigure}
\usepackage{mwe}

\usepackage{amsmath,amsfonts,bm}

\newcommand{\ours}{\textsc{Grape}\xspace}
\newcommand{\ourslong}{\underline{G}lobal \underline{R}egression with \underline{A}daptive \underline{P}rediction \underline{E}nhancement\xspace}

\def\eqref#1{equation~\ref{#1}}

\def\1{\bm{1}}

\def\vc{{\bm{c}}}

\def\vt{{\bm{t}}}

\DeclareMathAlphabet{\mathsfit}{\encodingdefault}{\sfdefault}{m}{sl}
\SetMathAlphabet{\mathsfit}{bold}{\encodingdefault}{\sfdefault}{bx}{n}

\def\sR{{\mathbb{R}}}

\newcommand{\mypar}[1]{\textbf{#1.}\hspace{4pt}}

\begin{document}
%

\title{Streaming Virtual Metrology via Lightweight Adaptation under Temporal Distribution Shifts
}

\author{\IEEEauthorblockN{Doyun Choi*}
\IEEEauthorblockA{
Seoul National University\\
Seoul, Republic of Korea \\
doyun.choi@snu.ac.kr}
\and
\IEEEauthorblockN{Cheonwoo Lee*}
\IEEEauthorblockA{
KAIST\\
Daejeon, Republic of Korea \\
cheonwoo.lee@kaist.ac.kr}
\and
\IEEEauthorblockN{Younghyun Han}
\IEEEauthorblockA{
Samsung Electronics\\
Hwaseong, Republic of Korea \\
yh92.han@samsung.com}
\and
\IEEEauthorblockN{Kiwook Han}
\IEEEauthorblockA{Samsung Electronics\\
Hwaseong, Republic of Korea \\
woog.han@samsung.com}
\and
\IEEEauthorblockN{Jinkook Park}
\IEEEauthorblockA{Samsung Electronics\\
Hwaseong, Republic of Korea \\
quip.park@samsung.com}
\and
\IEEEauthorblockN{Sungyoon Ryu}
\IEEEauthorblockA{Samsung Electronics\\
Hwaseong, Republic of Korea \\
sy625.ryu@samsung.com}
\and
\IEEEauthorblockN{Younghoon Sohn}
\IEEEauthorblockA{
Samsung Electronics\\
Hwaseong, Republic of Korea \\
yh.sohn@samsung.com}
\and
\IEEEauthorblockN{Jaemin Yoo}
\IEEEauthorblockA{
Seoul National University\\
Seoul, Republic of Korea \\
jaeminyoo@snu.ac.kr}
\thanks{*: These authors contributed equally. Author order is alphabetical.}
}

\maketitle


%


\maketitle

\begin{abstract}
Virtual metrology (VM) aims to estimate costly or delayed metrology variables from heterogeneous sensor signals and in-line measurements, and it acts as a critical component of semiconductor manufacturing.
While recent data-driven models have improved VM accuracy, most are trained in batch settings and assume stationary data distributions.
In practice, however, production environments evolve continuously due to equipment degradation, process adjustments, and maintenance events, making conventional batch-trained VM models difficult to maintain under temporal distribution shifts.
In this work, we propose \ours (\ourslong), a practical online learning framework for VM in evolving wafer-processing environments.
\ours integrates a stable global predictor based on interpretable representations with two lightweight streaming adapters that correct feature-level and environment-level shifts through online updates.
Experiments on two real-world semiconductor manufacturing datasets, including chronologically held-out blind-test streams, show that \ours improves robustness and generalization under 
real temporal drift compared with both conventional batch-learning methods and existing online learning baselines.
Code is available at https://github.com/cheonwoolee/GRAPE.
\end{abstract}

\begin{IEEEkeywords}
Time Series Regression, Online Learning, Distribution Shift, Semiconductor Manufacturing, Virtual Metrology
\end{IEEEkeywords}


%
\IEEEpeerreviewmaketitle
\section{Introduction}
\label{sec:intro}

Semiconductor manufacturing is a key industrial domain for data-driven modeling and decision support~\cite{ibrahim2025burnin,chen2025review,maitra2024virtual,fan2020data}.
As each semiconductor wafer moves through a sequence of processing steps, equipment records large volumes of sensor data, such as temperature, pressure, and gas flow.
These measurements provide critical information about process conditions and are closely related to whether the final chips meet quality goals.

In this context, virtual metrology (VM) uses readily available signals to predict quality variables that are expensive or delayed to measure, enabling faster feedback and tighter process control in production lines~\cite{han2025comparative,kang2009virtual,suthar2019featurevm}. 
As VM improves process monitoring and quality control, recent studies have developed machine learning- and deep learning-based VM approaches that train data-driven predictors on historical wafer data under static train-test protocols
~\cite{tin2022dlvm}.

Industrial VM data and modeling practices are characterized by three key properties.
First, the observations are highly heterogeneous, as they are collected from multiple sensing modalities with distinct statistical characteristics and  sampling behaviors. 
Second, the underlying data distributions are non-stationary, as operating conditions change due to equipment degradation, process adjustments, and maintenance events~\cite{cai2021jitpf,feng2019slowdriftvm,kang2017intelligentvm}.
Third, interpretability is crucial for VM deployment, as identifying target-relevant features helps engineers diagnose process behavior and make reliable decisions~\cite{mevic2025explainable}.

These properties pose significant challenges for reliable VM deployment in real semiconductor production environments. Heterogeneity makes it difficult to reliably detect distribution shifts consistently and to develop unified predictive models that remain robust across operating contexts~\cite{quinonero2008dataset,9847099}.
Non-stationarity further induces various forms of temporal shift, including covariate drift~\cite{gama2014survey,lu2019concept}, which changes the relationship between sensor features and metrology targets, and label shift~\cite{lipton2018labelshift}, which changes the distribution of the target variable itself.
Interpretability introduces another challenge, as industrial VM models should provide transparent prediction behavior that helps process engineers understand which sensor patterns and operating conditions drive each output. Limited interpretability makes it difficult to analyze process behavior and support reliable decision-making in production lines.


However, no existing approach addresses these three challenges at the same time.
Deep learning-based VM models can capture complex sensor patterns, but they often struggle with heterogeneous and irregular industrial VM data and provide limited interpretability for process diagnosis.
Existing online learning methods aim to address non-stationarity~\cite{vovk1997competitive, zabrocki2023adaptive}, but they typically update the predictive model uniformly and focus on a narrow range of temporal shifts, making it difficult to selectively respond to evolving factors while preserving stable process knowledge.

To address these challenges, we propose \ours (\ourslong), a practical online VM prediction framework that combines a stable global predictor with lightweight streaming adaptation.
First, we train a global prediction model to capture the general relationship between sensor observations and the target VM variable.
The model is built on interpretable features designed to represent diverse characteristics across heterogeneous sensor groups, making the prediction process more transparent for industrial analysis.
We then introduce two complementary adapters: (i) a feature-level adapter that tracks residual-driven changes in time-series behavior and (ii) an environment-level adapter that corrects context-specific offsets.
Together, these components support online learning through per-wafer updates, matching the sequential nature of VM deployment.
The staged design of \ours enables efficient adaptation to evolving production conditions without full retraining, while preserving the stability of the global predictor.

We conduct comprehensive experiments on two real semiconductor VM datasets under temporal drift.
Across multiple evaluation settings, including held-out test and unseen blind-test streams, \ours achieves the best overall average rank 
across representative offline and online baselines.
In particular, on the unseen blind-test streams, \ours achieves the best performance across all datasets and all three metrics,
with metric-averaged relative gains of 6.1\% and 10.1\% over the 
strongest competing methods on each dataset, respectively, highlighting its stronger adaptability under realistic deployment conditions.
We further present ablation studies and case-study analyses, demonstrating that the stable global backbone supports long-horizon generalization, while lightweight online adaptation recovers local regime changes without destabilizing predictions.
These results highlight the practical value of \ours for robust VM in evolving semiconductor production environments.

Our main contributions are summarized as follows:
\begin{itemize}
     \item \textbf{Industrial data insights:} We provide a systematic empirical characterization of real semiconductor VM data and demonstrate how temporal drift arises in industrial production processes, motivating the need for adaptive VM modeling under evolving operating conditions.
     \item \textbf{Practical modeling framework:} We propose \ours, a practical framework that combines a stable global backbone with two lightweight adapters for feature-level and environment-level shift correction. This design enables per-wafer adaptation without full model retraining, matching the sequential nature of VM deployment.
     \item \textbf{Real-world validation under temporal drift:} We validate \ours on two real semiconductor datasets, including held-out test and unseen blind-test streams. \ours achieves the best overall average rank across representative offline and online baselines.
     Our extensive ablation and case-study analyses provide deeper insight into its behavior under temporal drift.
\end{itemize}


\section{Problem Definition and Related Works}
\label{sec:related}

\subsection{Problem Definition}
Let $\mathcal{D} = \{(x_i,y_i)\}_{i=1}^{N}$ denote a 
dataset of wafer samples used for virtual metrology (VM), where $x_i$ represents the process observations of the $i$-th wafer and $y_i$ denotes the corresponding target metrology value. Here, $N$ denotes the total number of wafer samples.
For each wafer $i$, $x_i$ is represented as
\begin{equation}
    x_i = \{\mathbf{c}_i, \{\mathbf{o}_i^f, \mathbf{t}_i^f\}, \{\mathbf{o}_i^o, \mathbf{t}_i^o\}\},
\end{equation}
where $\mathbf{c}_i$ denotes the contextual identifiers of the wafer process (e.g., equipment, chamber, lot, and slot), which characterize the processing environment. The superscripts $f$ and $o$ correspond to fault detection and classification (FDC) and optical emission spectroscopy (OES) sensing data, respectively. 
For each sensing modality, the observations consist of a sequence of process time points $\mathbf{t}_i^* \in \mathbb{R}^{T_i^*}$ and sensing values collected from multiple sensors $\mathbf{o}_i^* \in \mathbb{R}^{S^* \times T_i^*}$.
Here, $T_i^*$ denotes the number of time points and $S^*$ denotes the number of sensors for modality *.
Given $x_i$, the goal of VM is to estimate the corresponding metrology value $y_i\in\sR$. 

Since wafers are processed sequentially over time, VM naturally operates in an online setting. At the moment of predicting a wafer, only past wafers and their observed metrology values are available, while future observations are unknown.
We therefore formulate VM as an online regression problem with history-dependent prediction.
Assume that wafer samples are chronologically ordered by processing time. For the $i$-th wafer, we define the historical context as
\begin{align}
    \mathcal{H}_i = \{(x_1,y_1),\ldots,(x_{i-1},y_{i-1})\}.
\end{align}
Then, the objective is to learn a regression model $f_{\theta_i}(\cdot)$ that predicts $y_i$:
\begin{equation}
\hat{y}_i = f_{\theta_i}(x_i, \mathcal{H}_i),
\end{equation}
where $\theta_i$ denotes the model parameters updated using the historical observations $\mathcal{H}_i$ available prior to predicting $y_i$.

Considering the time-varying nature of the VM deployment environment, we adopt a chronological evaluation setup that reflects the real-world production settings.
The first $N_t$ wafers are used as the training set $\mathcal{D}_t=\{(x_i,y_i)\}_{i=1}^{N_t}$ with $N_t < N$, and the remaining wafers are processed sequentially in temporal order for validation and testing.
Once the prediction for a wafer is made, its ground-truth value becomes available for subsequent model updates. This protocol prevents information leakage from future wafers and enables the assessment of VM performance under realistic temporal distribution shifts.

\subsection{Online Learning}


Online learning is a paradigm in which 
models are updated sequentially as new observations arrive, rather than being trained once on a fixed historical dataset~\cite{mcmahan2017survey,NEURIPS2021_5e6bd7a6}. This paradigm is well suited to non-stationary environments, where the data-generating process changes over time~\cite{guo2025online}. In such settings, models trained only on historical data can gradually become mismatched to current operating conditions, leading to degraded predictive performance. By incrementally updating model parameters using newly observed samples, online learning enables continuous adaptation without repeated retraining.

In VM, online learning has been studied as a practical approach to handling distribution shifts in semiconductor processes. For example, online linear models update predictors sequentially using prediction feedback~\cite{vovk1997competitive}, while AdaOTP incorporates time-aware normalization and temporal ensemble prediction to adapt to evolving data streams~\cite{zabrocki2023adaptive}.
Nevertheless, applying online learning to VM remains fundamentally challenging, because semiconductor process data are highly heterogeneous, irregular, temporally drifting, and often require interpretable predictions for reliable decision-making, as discussed in Section~\ref{sec:obs}.

\subsection{Virtual Metrology (VM)}

VM aims to estimate product quality or metrology variables using process data collected during manufacturing, reducing the need for costly and time-consuming physical measurements. In semiconductor fabrication, physical metrology is typically performed only on sampled wafers, which limits real-time monitoring and process control. VM models
address this limitation by predicting metrology values, such as film thickness, critical dimension, or etch depth, from equipment sensors and process parameters~\cite{Dreyfus17012022}.
Early VM studies mainly employed statistical models, including partial least squares regression, support vector regression, and Gaussian process regression~\cite{CHEN2020192}. Representative studies have also shown that VM predictions can be integrated into run-to-run control frameworks~\cite{KANG20112508}.

Recent research has explored more advanced machine learning approaches to address the nonlinear and high-dimensional nature of manufacturing sensor data. Ensemble learning methods and deep learning architectures have been applied to capture complex relationships between process signals and wafer characteristics~\cite{CHEN2020192,maitra2024virtual}. In particular, deep neural networks can automatically extract informative representations from large-scale sensor data such as fault detection and classification (FDC) signals or optical emission spectroscopy (OES), improving prediction performance compared to traditional feature-engineering approaches~\cite{maitra2024virtual}. Despite these advances, VM modeling remains challenging due to limited labeled metrology data, distribution shifts caused by process drift, and the need for models that generalize across tools and process conditions~\cite{maitra2024virtual}.

\begin{figure}[t]
    \centering
    \subfigure[Heterogeneity]{
        \includegraphics[width=0.21\textwidth]{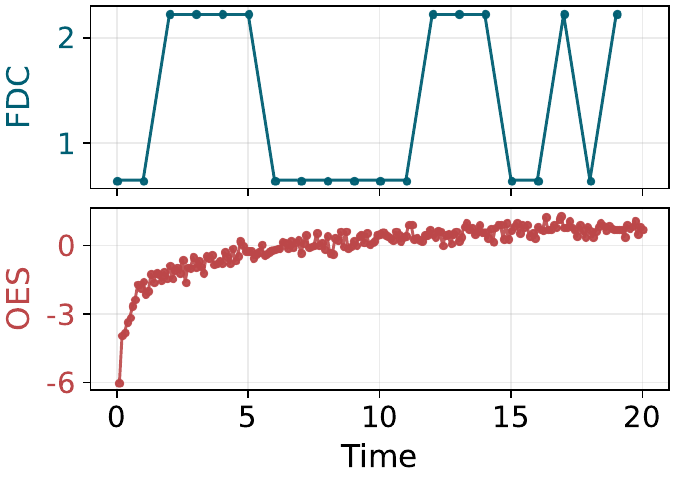}
        \label{fig:data-hetero}
    }\hfill
    \subfigure[Irregularity]{
        \includegraphics[width=0.21\textwidth]{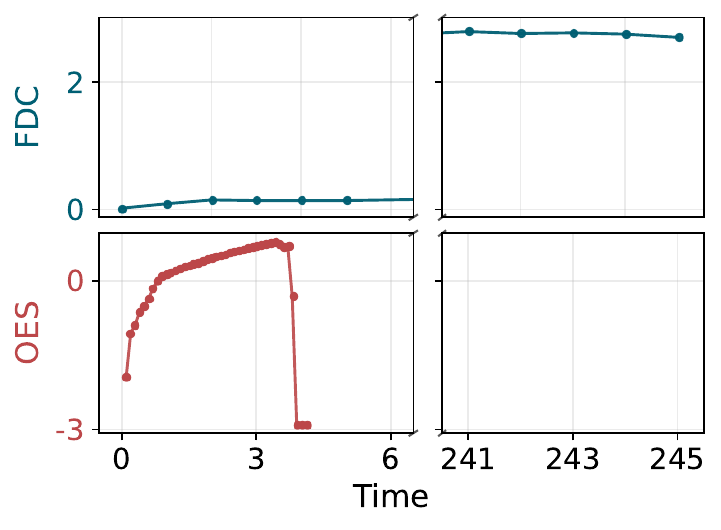}
        \label{fig:data-irr}
    }
    \caption{
    Distinct properties of industrial VM data.
    (a) Heterogeneity: different sensing modalities exhibit distinct signal patterns and sampling rates.
    (b) Irregularity: sensor observations are recorded at non-uniform time intervals across sensors.}
    \label{fig:hetero-and-irr}
\end{figure}

\section{Challenges in Industrial VM}
\label{sec:obs}

In semiconductor manufacturing, virtual metrology (VM) models are developed on diverse and heterogeneous data sources. 
We summarize key challenges observed in real-world industrial VM data and modeling practice.


\subsection{Heterogeneity and Irregularity (C1)}
\label{ssec:c1}

Industrial VM data are both heterogeneous and irregular, which 
distinguishes them from conventional multivariate time series. 
For each wafer sample $x_i$, the observations $\mathbf{o}_i^*$ are collected from multiple heterogeneous sensors in two groups: FDC and OES.
As illustrated in Fig.~\ref{fig:hetero-and-irr}, these sensing modalities reflect different physical aspects of wafer processing and operate at different sampling frequencies and measurement intervals.
In particular, as shown in Fig.~\ref{fig:hetero-and-irr}(b), long periods without observations can occur, further increasing the irregularity of the time series.
As a result, the associated time sequences $\mathbf{t}_i^*$ and sensor observations $\mathbf{o}_i^*$ form highly irregular multivariate time series.


\begin{figure}[t]
    \centering
    \subfigure[Covariate shift]{
        \includegraphics[width=0.21\textwidth]{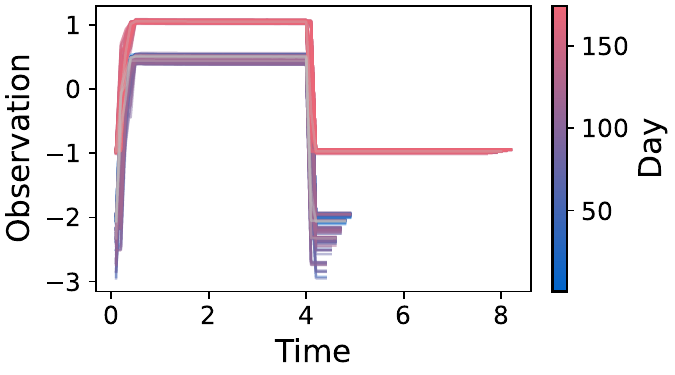}
        \label{fig:cov-shift}
    }\hfill
    \subfigure[Label shift]{
        \includegraphics[width=0.21\textwidth]{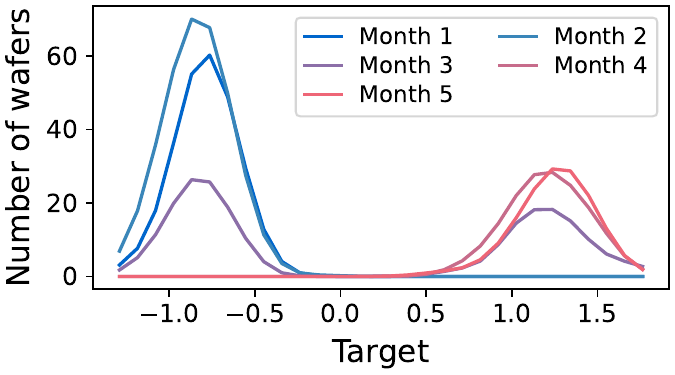}
        \label{fig:label-shift}
    }
    \caption{Representative temporal distribution shifts in industrial VM data. 
    (a) The trajectories of representative sensors, colored by recording time, show substantial covariate shift. 
    (b) Distribution of prediction targets by recording month shifts toward larger values over time, illustrating temporal label shift.}
    \label{fig:X-shift}
\end{figure}

\begin{figure*}[t]
\centering
\includegraphics[width=\textwidth,trim={0.3cm 0.1cm 0.4cm 0.1cm},clip]{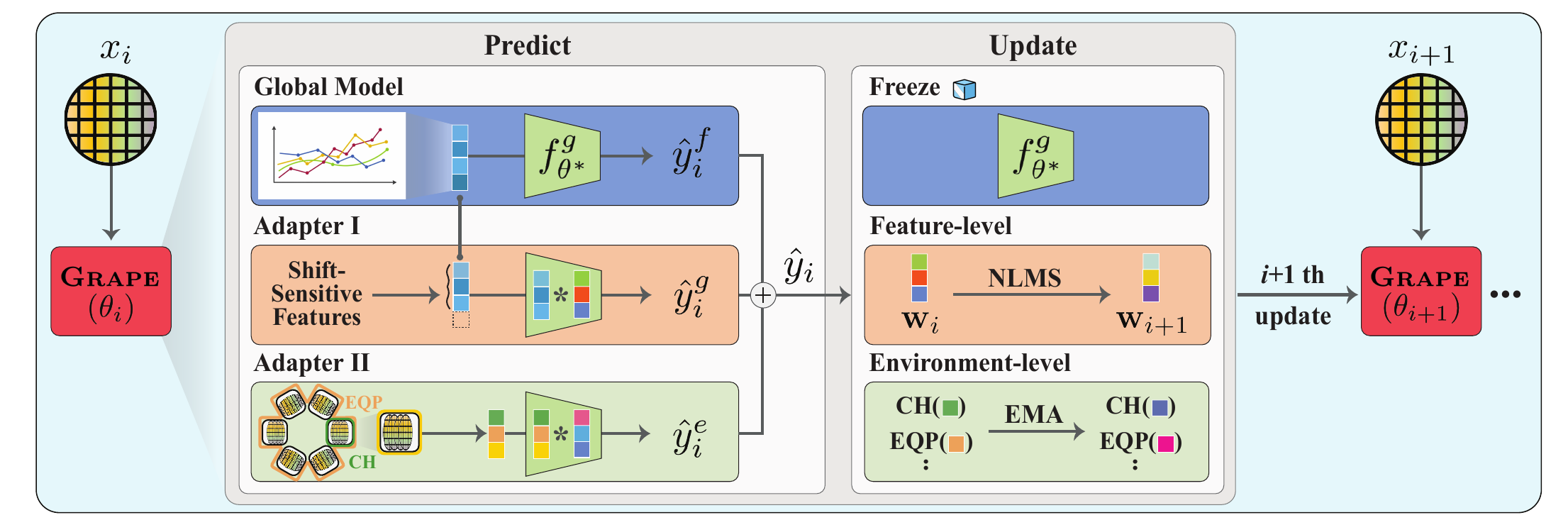}

\caption{Overall framework of \ours. For each wafer $i$, the global model $f_{\theta^*}^{g}$ produces an initial prediction based on target-aware representations. Adapter~I performs feature-level adaptation by adjusting the contributions of shift-sensitive features, while Adapter~II corrects prediction offsets at the environment level. The final prediction is obtained by combining the outputs of these components, and the adapters are updated online to continuously adapt the model to temporal distribution shifts.}
\label{fig:overallframework}
\end{figure*}




\subsection{Temporal Distribution Shift (C2)}
\label{ssec:c2}

Temporal distribution shift is a well-known challenge in VM data, often appearing as gradual data drift caused by equipment aging or as abrupt data shifts caused by process adjustments, maintenance events, and sensor calibration drift~\cite{cho2025physical,zabrocki2023adaptive}.
These shifts can steadily degrade model accuracy or cause abrupt prediction failures, making static VM models unreliable in evolving production environments.
In this work, we observe two major forms of temporal distribution shift in real-world industrial VM data: (i) covariate shift and (ii) label shift.

\emph{Covariate shift} refers to a scenario where the input distribution $p(x)$ gradually changes over time~\cite{FU2026122618,raza2016adaptive}. 
As illustrated in Fig.~\ref{fig:cov-shift}, representative sensor trajectories exhibit a gradual increase in observation levels over processing time, indicating that wafer observations gradually shift toward different value ranges.
Such changes may appear as temporal offsets along the time axis (i.e., shifts in $\vt_i$) or as changes in equipment conditions (i.e., shifts in $\vc_i$).
These variations alter the statistical properties of sensor observations, reducing the reliability of models trained on historical data.

Beyond covariate shift, \emph{label shift} occurs when the label distribution $p(y)$ changes over time~\cite{lipton2018labelshift}. 
As shown in Fig.~\ref{fig:label-shift}, the target distribution changes noticeably across months, with its dominant mass moving from lower values in earlier periods to higher values in later periods.
This distributional change poses a significant challenge for predictive models, as the learned model may no longer accurately reflect the behavior of future wafers.



\subsection{Lack of Interpretability (C3)}
\label{ssec:c3}

Effective VM modeling often depends on identifying features that are strongly related to the target variable. However, such feature design typically requires substantial domain expertise, which is costly and difficult to obtain in semiconductor manufacturing due to strict industrial confidentiality. Interpretability is therefore essential, as VM predictions are expected to support not only quality estimation but also process diagnosis and operational decision-making.

Nevertheless, many existing approaches rely on deep learning-based encoders to automatically learn representations from sensor data. While these methods can improve predictive accuracy, they often behave as black-box models, offering little explanation of which sensors or signal patterns influence the output. Consequently, engineers find it difficult to understand 
model behavior, verify prediction reliability, or relate changes in predictions to underlying process variations.







\section{Proposed Method}
\label{sec:method}

We propose \ours (\ourslong), an online learning framework designed to address the aforementioned challenges in virtual metrology (VM).
Our idea is to decouple stable global modeling from adaptive online correction, enabling the model to capture general process relationships while dynamically adapting to evolving production environments.

\subsection{Overview}

Fig.~\ref{fig:overallframework} presents the overall architecture of \ours, which is designed to address the three challenges described above.
First, to address \textbf{C1} (\emph{heterogeneous and irregular VM data}), we train a global prediction model $f_\theta^{g}$ that captures the general relationship between wafer process signals and target metrology values. This model operates on target-aware representations produced by a feature generator, modeling heterogeneous and irregular observations (\S\ref{ssec:globlmodel}).

Second, to address \textbf{C2} (\emph{temporal distribution shift}), we utilize two lightweight adapters that adjust the global model from complementary perspectives during deployment. One adapter accounts for feature-level shifts by adjusting the contribution of shift-sensitive features, while the other corrects prediction offsets using recent environment-level residual statistics. Together, these  adapters allow the model to remain robust under evolving process conditions (\S\ref{ssec:onlinelearning}).

Third, to address \textbf{C3} (\emph{lack of interpretability}), we adopt an additive prediction formulation in the inference and update stages. This design decomposes the final prediction into explicit components associated with the global predictor and the two adaptation modules, making the effect of each part transparent and easier to interpret. The online update procedure further enables practitioners to track how predictions change over time for newly observed data (\S\ref{ssec:inf-and-upd}).

\subsection{Pre-training the Global Model}
\label{ssec:globlmodel}

A key component of \ours is the global prediction model, which captures the general relationship between wafer process observations and the target metrology value.
The global model $f_\theta^g$ is trained using the training dataset $\mathcal{D}_t$,
and its parameters $\theta$ remain fixed during evaluation.

A key challenge lies in constructing informative representations from the wafer observations $x_i$, as the heterogeneity and irregularity of VM data make representation learning difficult to apply directly.
Accordingly, we design a target-aware feature constructor that extracts interpretable and task-aware features from wafer observations. 
Specifically, to address heterogeneous sensor modalities, we extract diverse signal descriptors. To handle irregular temporal observations, we transform raw time-series signals into feature-level summaries. 

\begin{table}[t]
\centering
\caption{Target-aware feature groups derived from wafer sensing signals.}
\label{tab:feature_groups}
\setlength{\tabcolsep}{10pt}
\resizebox{\columnwidth}{!}{
\begin{tabular}{ll}
\toprule
\textbf{Feature Group} & \textbf{Description} \\
\midrule
Global/window statistics 
& Mean, min, max, std of sensor signals. \\

Temporal dynamics 
& Rising time, peak time, settling time. \\

Shape characteristics 
& Max gradient, monotonicity, curvature. \\

Frequency-domain features 
& Dominant frequency, top FFT magnitudes. \\

Cross-sensor correlations
& Mean correlation across sensor groups. \\

Window transitions 
& Changes in sensor statistics across windows. \\
\bottomrule
\end{tabular}
}
\end{table}

Table~\ref{tab:feature_groups} summarizes the six groups of features constructed from wafer sensing signals. These features characterize different aspects of wafer process signals from multiple perspectives, allowing the representation to capture diverse properties associated with heterogeneous sensing modalities. 
Given the wafer observation $x_i$, we generate candidate features $\mathbf{z}_i\in\mathbb{R}^F$, where $F$ is the total number of candidate features.

Among these candidate features, only a subset may be informative for predicting the target variable. To identify persistent and target-relevant signals, we select features based on their correlations with the target variable $y_i$.
Specifically, we compute the Pearson correlation coefficient~\cite{benesty2009pearson} between each candidate feature $z_{i, f}$ and $y_i$ across the training wafers:
\begin{equation}
\label{eq:PCC}
\rho_f =
\frac{
\sum_{i=1}^{N_t}
(z_{i,f} - \bar{z}_f)(y_i - \bar{y})
}{
\sqrt{\sum_{i=1}^{N_t}(z_{i,f} - \bar{z}_f)^2}
\sqrt{\sum_{i=1}^{N_t}(y_i - \bar{y})^2}
}.
\end{equation}
Here, $z_{i,f}$ denotes the value of the $f$-th candidate feature for wafer $x_i$, $\bar{z}_f$ represents the mean of $z_f$, and $\bar{y}$ denotes the mean of $y_i$.

We then select the top-$F_g$ features with the largest absolute correlation values $|\rho_f|$, where $F_g$ is a hyperparameter controlling the dimensionality of the feature space.
The selected feature set is denoted by $\mathcal{F}_g$, where $|\mathcal{F}_g| = F_g$. 
The selected features are concatenated to form a $F_g$-dimensional feature vector $\mathbf{z}_{i,\mathcal{F}_g} \in \mathbb{R}^{F_g}$ for each wafer sample $x_i$. This vector serves as the target-aware representation used to train the prediction model.
Finally, we normalize each feature dimension to maintain a consistent scale and to prevent features with larger magnitudes from dominating the learning process.

Using the feature representation $\mathbf{z}_{i,\mathcal{F}_g}$, we train a global prediction model $f_\theta^g$  by minimizing the mean squared error (MSE) over the training dataset:
\begin{equation}
\mathcal{L}_{\text{MSE}} =
\frac{1}{N_t}
{\textstyle \sum_{(x_i,y_i)\in\mathcal{D}_{t}}}
(y_i-f_\theta^g(\mathbf{z}_{i,\mathcal{F}_g}))^2.
\end{equation}
After its training, $f_{\theta^*}^{g}$ generates baseline predictions and remains fixed during the online-learning stage.
In principle, $f_\theta^{g}$ can be implemented using any regression model, such as linear regression or a multilayer perceptron, which makes a different tradeoff between performance and stability.
In our implementation, we adopt Ridge regression for stable predictions and improved generalization.

\subsection{Streaming Adaptation to Distribution Shifts}
\label{ssec:onlinelearning}

As discussed in Section~\ref{ssec:c2}, data distributions in wafer manufacturing processes often evolve over time. As a result, a fixed global prediction model may gradually degrade in performance when applied to future wafer samples.

We therefore introduce an adaptation algorithm that enables the model to respond to temporal distribution shifts. 
During deployment, wafer samples arrive sequentially, and the VM predictor processes each wafer in a streaming manner.
To adapt predictions to distribution shifts, we introduce two adapter modules that refine the outputs of the global model from complementary perspectives. 

\mypar{Adapter I. Modeling Feature-Level Distribution Shifts}
This component $f_{\phi_i}^f$ focuses on distribution shifts in the sensing observations $\mathbf{o}_i^*$.
The motivation is that if the contribution of some features changes over time, the global model will exhibit systematic residuals that are correlated with those features.
We therefore hypothesize that features strongly correlated with prediction residuals are more susceptible to temporal distribution shifts.

First, we compute the residual $\mathbf{e}\in\mathbb{R}^{N_t}$ on the training data, where the $i$-th element represents the prediction error of the global model for sample $x_i$:
\begin{equation}
e_i = y_i - f_{\theta^*}^{g}(\mathbf{z}_{i,\mathcal{F}_g}).
\end{equation}

We then identify features that are highly correlated with the residual errors using the Pearson correlation coefficient. Based on the absolute correlation values, we select the top-$F_a$ features, where $F_a$ is a hyperparameter. Let $\mathcal{F}_a\subseteq\mathcal{F}_g$ denote the selected subset.
Using this subset, we construct the feature vector $\mathbf{z}_{i,\mathcal{F}_a}$ for each streamed wafer sample and compute the adapter output as
\begin{equation}
\hat{y}_{i}^f = f_{\phi_i}^f(\mathbf{z}_{i,\mathcal{F}_a})=
(\mathbf{w}_i^{f})^\top
\mathbf{z}_{i,\mathcal{F}_a},
\end{equation}
where $\mathbf{w}_i^f \in \mathbb{R}^{F_a}$ denotes the adapter weight vector. The weights are initialized as $\phi_0=\mathbf{w}_0^f=\mathbf{0}$ to ensure that the first update is driven solely by the observed error signal without any bias from the parameter initialization. The weights are subsequently updated during streaming inference.

\mypar{Adapter II. Modeling Environment-Level Distribution Shifts}
While Adapter I captures feature-level distribution shifts, changes may also arise from processing environments $\mathbf{c}_i$. To address this issue, we introduce a second adapter $f_{\psi_i}^e$ that models environment-level shifts using hierarchical residual correction.

The motivation is that wafers processed under the same equipment or chamber conditions often share similar prediction biases, even when such effects are not fully explained by the input features.
Accordingly, we exploit the hierarchical structure of wafer processing to adapt predictions under distribution shifts.
Based on the contextual identifier $\mathbf{c}_i$ associated with wafer $x_i$, we track residual statistics across a hierarchy set $\mathcal{L}$. In our setting, $\mathcal{L}$ contains four levels: global, equipment, chamber, and equipment-chamber pair.
The correction is computed as a weighted combination of residual means across the hierarchy levels:
\begin{equation}
\hat{y}_{i}^e = f_{\psi_i}^e(\mathbf{c}_i)={\textstyle \sum_{l\in \mathcal{L}(\mathbf{c}_i)}}  \tilde{\omega}_i^{l} \mu_i^l + \delta_{i},
\end{equation}
where $\mathcal{L}(\mathbf{c}_i)$ denotes the set of available hierarchy levels for the context $\mathbf{c}_i$, $\mu_i^l$ is the residual mean at level $l$, and $\tilde{\omega}_i^{l}$ is the normalized reliability weight obtained by normalizing $\omega_i^{l}$ across all hierarchy levels.
The extreme-shift correction term $\delta_{i}$ provides an additional correction that is activated when extreme distribution shifts exceed a predefined threshold.
The initial values are computed as
\begin{equation}
\displaystyle \omega_0^l = \frac{n_0^l}{n_0^l + n^{\min}}, \quad \mu_{0}^l = \frac{1}{n_0^l}
{\textstyle \sum_{i \in \mathcal{L}(\mathbf{c}_i)}} (y_i-\hat{y}_i^g),\quad \delta_{0}=0,
\end{equation}
where $n_0^l$ is the number of observations associated with level $l$ in the training dataset and $n^{\min}$ is a hyperparameter controlling the shrinkage strength.


\subsection{Interpretable Inference and Updates}
\label{ssec:inf-and-upd}

The final prediction of the proposed framework is computed as the sum of the outputs from the global model and the two adapters:
\begin{equation}
\label{eq:finalprediction}
\hat{y}_i = f_{\theta_i}(x_i, \mathcal{H}_i)= \hat{y}_{i}^g
+
\hat{y}_{i}^f
+
\hat{y}_{i}^e,
\end{equation}
where $\Theta=\{\theta^*, \psi_i, \phi_i\}$.
This additive formulation provides a transparent correction structure: $\hat{y}_{i}^{g}$ is maintained as the baseline prediction, while $\hat{y}_{i}^{f}$ and $\hat{y}_{i}^{e}$ act as adaptive correction terms.
After obtaining the prediction $\hat{y}_i$, the adapters are updated in a streaming manner using newly observed data, allowing the system to continuously adapt to evolving process conditions.

For adapter I, the weights are updated online using the prediction residual of the current wafer sample. Specifically, the update follows the normalized least mean squares (NLMS) rule~\cite{1433141}, which provides a simple and computationally efficient approximation of stochastic least-squares updates:
\begin{equation}
\mathbf{w}_{i+1}^{f} =
\mathbf{w}_{i}^{f}
+
\eta
\frac{e_i \mathbf{z}_{i,\mathcal{F}_a}}
{\|\mathbf{z}_{i,\mathcal{F}_a}\|^2 + \lambda},
\end{equation}
where $e_i = y_i - \hat{y}_i$ denotes the prediction residual, $\eta$ is the learning rate controlling the update magnitude, and $\lambda$ is a small regularization constant. 
Intuitively, the adapter weights provide an interpretable indication of feature contributions: when large prediction residuals occur, the weights associated with shift-sensitive features are adaptively adjusted.


For adapter II, we maintain an exponentially weighted moving average (EMA) of the residuals and the number of observed samples. EMA provides a lightweight mechanism for tracking gradually evolving residual statistics while emphasizing recent observations. For each $l\in \mathcal{L}(\mathbf{c}_i)$, the parameters are updated as
\begin{equation}
\begin{aligned}
\omega_{i+1}^l
&= \frac{n_{i+1}^l}{n_{i+1}^l + n^{\min}}, \\
\mu_{i+1}^l
&= (1-\alpha)\mu_i^l
+ \alpha \left(y_i-\hat{y}_i^g-\hat{y}_i^f\right), \\
n_{i+1}^l
&= n_i^l + 1.
\end{aligned}
\end{equation}
where $\alpha$ is the smoothing constant that controls the influence of recent residual observations on the maintained statistics. Small fluctuations are smoothed by the EMA, while persistent prediction bias gradually shifts, enabling the adapter to correct systematic environment-level offsets in an interpretable manner. 

The correction term $\delta_{i+1}=\delta_i + \beta (y_i-\hat{y}_i)$ tracks the global residual $y_i-\hat{y}_i$ across streamed samples using a smoothed residual update
with a relatively large smoothing factor to detect abrupt distribution shifts.
By emphasizing recent residual observations, it helps capture extreme shifts that may not be immediately reflected in the smoother updates of Adapter~II. We set $\beta$ to $0.5$, while $\alpha$ is a hyperparameter mostly tuned in a smaller range.

\subsection{Computational Complexity of \ours}

Most of the computational cost of \ours is incurred during the pre-training stage, while online inference and adaptation remain lightweight during deployment. Although pre-training introduces super-linear computational complexity, this cost is incurred only once during initialization and does not affect per-wafer deployment efficiency.

For pre-training, \ours first standardizes the feature matrix, which is $O(N_tF)$ for $N_t$ training wafers and $F$ candidate features.
Ranking features according to their Pearson correlation with the target is $O(F\log F)$. 
The global predictor is trained using ridge regression, which is $O(N_tF_g^2)$ for computing the covariance terms among the $F_g$ selected features and $O(F_g^3)$ for the matrix inversion.
Adapter feature selection computes residual-feature correlations over the selected feature matrix, costing $O(N_tF_g + F_g\log F_g)$. Therefore, the overall time complexity is
$O(N_tF + F\log F + N_tF_g^2 + F_g^3 + N_tF_g + F_g\log F_g)$.
For memory requirements, it is $O(N_tF + F_g^2)$ for storing the candidate feature matrix and ridge-regression matrices.
In practice, these costs remain modest because the number of selected features $F_g$ is substantially smaller than the original feature dimension $F$ ($F_g\ll F$).

For each streamed wafer, \ours requires only lightweight computations.
The fixed global predictor computes a ridge-regression output in $O(F_g)$ time, where $F_g$ denotes the number of selected features used by the global model $f_{\theta^*}^g$.
Adapter I computes the feature-level correction in $O(F_a)$ time, where $F_a$ is the number of shift-sensitive features used by the adapter.
Adapter II computes the environment-level correction using a constant number of hash-table lookups over the predefined context hierarchy, resulting in $O(1)$ time complexity.
Therefore, the total per-wafer inference complexity is $O(F_g + F_a)$. The online update is also lightweight: Adapter I performs an $O(F_a)$ NLMS update over the selected features, while Adapter II updates EMA-based context statistics and the extreme-shift correction term in $O(1)$ time.

The memory overhead of online adaptation is also effectively bounded.
Adapter I maintains an $F_a$-dimensional weight vector, requiring $O(F_a)$ memory.
Adapter II stores a small set of context-level EMA statistics and the extreme-shift correction term for the fixed hierarchy of environment identifiers.
Including the fixed global model weights, the total memory requirement is $O(F_g + F_a)$.

These complexity properties make \ours suitable for practical VM deployment in evolving semiconductor manufacturing environments.
Since each wafer can be processed and updated with linear complexity in the selected feature dimensions and without full model retraining, \ours provides a computationally efficient mechanism for maintaining robust predictions in streaming production lines.


\section{Experiments}
\label{sec:exp}

We conduct a comprehensive evaluation on two real-world semiconductor virtual metrology (VM) datasets to assess the robustness of \ours under temporal distribution shifts. We further conduct blind-test, ablation, and case-studies 
to provide a deeper understanding of model behavior.

\mypar{Datasets}
We conduct experiments on two real-world industrial VM datasets, \textbf{D1} and \textbf{D2}, collected from an etch process, with the goal of predicting the optical critical dimension.
\textbf{D1} contains 899 wafer samples, while \textbf{D2} has 5,798 samples.
In both datasets, each wafer is represented by multivariate time series observations from multiple sensor groups, including FDC and OES.
For preprocessing, we retain only observations that are commonly available in both data sources, and exclude sensors with more than 95\% missing values across all observations.
The datasets exhibit temporal shifts in both the input and target spaces, corresponding to covariate shift in $x$ and label shift in $y$, respectively.
Due to confidentiality issues in semiconductor manufacturing, all dataset identifiers and feature names are anonymized. 

\mypar{Baselines}
We compare \ours with seven strong baselines, covering both batch and online learning methods. We first consider four batch-based offline methods:
\begin{itemize}

\item \textbf{Linear.}
A standard linear regression model trained on the historical training set using ordinary least squares. We employ ridge regression with $\ell_2$ regularization. 

\item \textbf{MLP.}
A multilayer perceptron consisting of fully connected layers with nonlinear activations, enabling it to capture complex relationships between input features and the target variable.

\item \textbf{DLinear~\cite{zeng2023transformers}.}
It is a lightweight architecture designed for time-series forecasting.
It decomposes the input sequence into trend and seasonal components and applies linear layers to each component separately.

\item \textbf{Transformer~\cite{NIPS2017_3f5ee243}.}
A standard Transformer-based sequence model that captures temporal dependencies using self-attention.
It enables flexible modeling of long-range temporal interactions.
\end{itemize}

\begin{table}[t]
\caption{Test performance on two semiconductor VM datasets.
\ours achieves the best average rank across various metrics and datasets, demonstrating improved robustness compared with both batch and online learning approaches.}
\label{tab:test-performance}
\centering
\large
\resizebox{\columnwidth}{!}{
\setlength{\tabcolsep}{6pt}
\renewcommand{\arraystretch}{1.05}
\begin{tabular}{l|ccc|ccc|c}
\toprule
\multirow{2}{*}{Method} 
& \multicolumn{3}{c|}{D1} 
& \multicolumn{3}{c|}{D2} 
& \multirow{2}{*}{\begin{tabular}{c}Avg. \\ Rank\end{tabular}} \\

\cmidrule(lr){2-4}\cmidrule(lr){5-7}

& RMSE & MAE & $R^2$ & RMSE & MAE & $R^2$ & \\

\midrule
\multicolumn{8}{c}{\textsc{Batch Methods}} \\
\midrule

Linear & 0.1712 & 0.1293 & 0.1256 & 1.0920 & 0.8980 & -0.4827 & 5.67 \\
MLP & 0.1715 & 0.1271 & 0.1231 & 1.0483 & 0.8532 & -0.3664 & 5.33 \\
DLinear & 0.1860 & 0.1441 & -0.0311 & 0.9313 & 0.7390 & -0.0783 & 6.50 \\
Transformer & 0.1823 & 0.1389 & 0.0095 & 0.8784 & 0.6840 & 0.0406 & 5.17 \\

\midrule
\multicolumn{8}{c}{\textsc{Online Learning Methods}} \\
\midrule

Rolling OLS & 0.1947 & 0.1514 & -0.1308 & 0.8860 & 0.6602 & 0.0240 & 6.17 \\
Online Linear & \underline{0.1652} & \textbf{0.1229} & \underline{0.1858} & 0.8525 & 0.6775 & 0.0964 & \underline{2.50} \\
AdaOTP & 0.1792 & 0.1358 & 0.0423 & \underline{0.7838} & \underline{0.6218} & \underline{0.2363} & 3.50 \\

\midrule
\textbf{\ours (ours)} & \textbf{0.1647} & \underline{0.1270} & \textbf{0.1910} & \textbf{0.7658} & \textbf{0.5990} & \textbf{0.2709} & \textbf{1.17} \\

\bottomrule
\end{tabular}
}
\end{table}

\begin{table}[t]
\caption{Performance under a blind-test setting on two semiconductor VM datasets.
\ours achieves the best performance, indicating improved generalization.
Batch methods degrade substantially, while online learning methods remain markedly stronger.
}
\label{tab:unseen-performance}
\centering
\large
\resizebox{\columnwidth}{!}{
\setlength{\tabcolsep}{6pt}
\renewcommand{\arraystretch}{1.05}
\begin{tabular}{l|ccc|ccc|c}
\toprule
\multirow{2}{*}{Method} 
& \multicolumn{3}{c|}{D1} 
& \multicolumn{3}{c|}{D2} 
& \multirow{2}{*}{\begin{tabular}{c}Avg \\ Rank\end{tabular}} \\

\cmidrule(lr){2-4}\cmidrule(lr){5-7}

& RMSE & MAE & $R^2$ & RMSE & MAE & $R^2$ & \\

\midrule
\multicolumn{8}{c}{\textsc{Batch Methods}} \\
\midrule

Linear & 1.2968 & 0.9080 & -0.5966 & 1.1381 & 0.9469 & -0.6078 & 6.50 \\
MLP & 1.2924 & 0.9076 & -0.5858 & 1.1511 & 0.9599 & -0.6447 & 6.50 \\
DLinear & 1.3245 & 0.9394 & -0.6612 & 0.9828 & 0.7893 & -0.2313 & 7.00 \\
Transformer & 1.3015 & 0.9223 & -0.6041 & 0.8689 & 0.6854 & 0.0377 & 6.00 \\

\midrule
\multicolumn{8}{c}{\textsc{Online Learning Methods}} \\
\midrule

Rolling OLS & 0.7281 & 0.4221 & 0.4968 & 0.8543 & 0.6606 & 0.0697 & 3.50 \\
Online Linear & 0.9882 & 0.6788 & 0.0729 & \underline{0.8097} & 0.6475 & \underline{0.1642} & 3.17 \\
AdaOTP & \underline{0.2314} & \underline{0.1677} & \underline{0.9492} & 0.8150 & \underline{0.6432} & 0.1534 & \underline{2.33} \\

\midrule
\textbf{\ours (ours)} & \textbf{0.2099} & \textbf{0.1540} & \textbf{0.9582} & \textbf{0.7962} & \textbf{0.6165} & \textbf{0.2132} & \textbf{1.00} \\

\bottomrule
\end{tabular}
}
\end{table}

We further compare \ours with three online learning methods previously proposed for VM:
\begin{itemize}
\item \textbf{Rolling OLS.} 
It refits an ordinary least squares estimator using the most recent observations within a fixed context window.
It is included as an online learning baseline for VM, following the experimental setting of~\cite{zabrocki2023adaptive}.

\item \textbf{Online Linear~\cite{vovk1997competitive}.}
It incrementally updates model parameters as new samples arrive in a streaming manner. A linear predictor adjusts its weights through incremental updates to minimize instantaneous prediction error.

\item \textbf{AdaOTP~\cite{zabrocki2023adaptive}.}
It is an adaptive prediction framework that combines multiple temporal predictors and dynamically adjusts their contributions based on per-sample losses. 
It incorporates time-aware normalization to account for temporal variation in the input signals.

\end{itemize}

For a fair comparison, Linear, MLP, and the online learning baselines use the same $F_g$-dimensional target-aware representation as the global model of \ours.
For the two end-to-end time-series regression baselines, DLinear and Transformer, we construct regular time-series inputs from irregular sensor observations using forward filling, where missing values are imputed with the most recent available observation.

\begin{figure*}[t]
    \centering
    \subfigure[An abrupt temporal shift occurs; \ours adapts online and gradually tracks the new target regime, while static baselines remain near pre-shift levels.
    While Rolling OLS and Online Linear adapt to the shift, they exhibit highly fluctuating predictions.
    ]{
        \includegraphics[width=\textwidth]{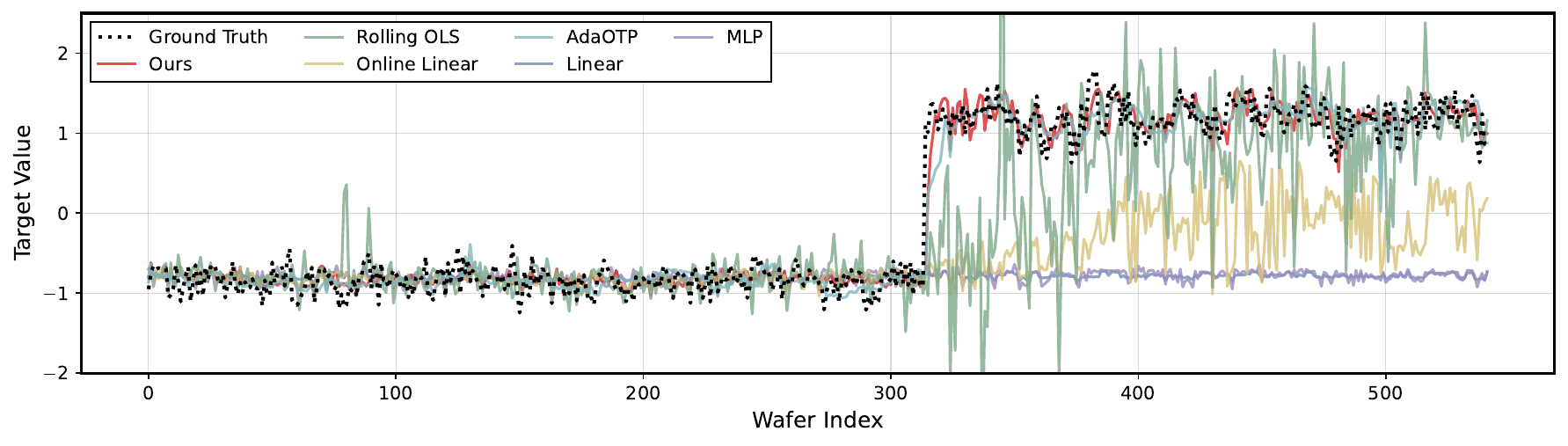}
        \label{fig:case-study-d1}
    }
    \subfigure[After the temporal shift takes place, \ours progressively adjusts its predictions to follow the evolving target values, whereas other methods fail to adapt.]{
        \includegraphics[width=\textwidth]{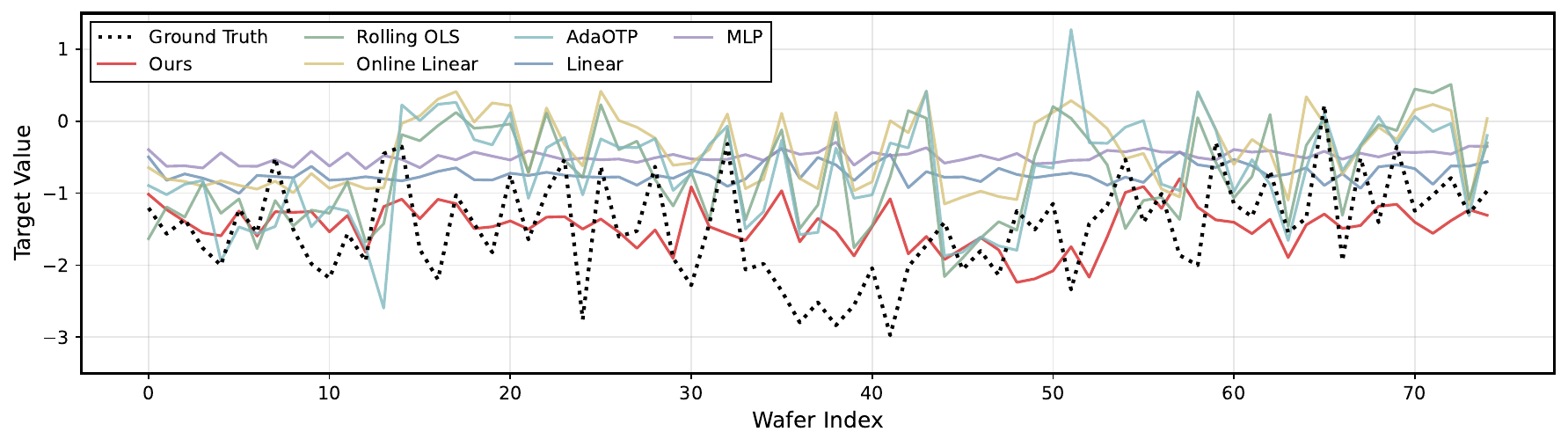}
        \label{fig:case-study-d2}
    }
    \caption{Case-study examples 
    from data streams under temporal shift: (a) D1 and (b) D2.}
    \label{fig:case-study}
\end{figure*}





\mypar{Experimental Settings}
For each dataset, wafers are chronologically ordered and partitioned into train, validation, test, and unseen splits with a 25:5:10:60 ratio.
To better reflect realistic deployment, the chronologically latest unseen split is used exclusively for blind-test evaluation and is never accessed during model selection or hyperparameter tuning.
For all methods, hyperparameters are selected via grid search based on validation $R^2$ and are reported on both the test and unseen splits.
All reported RMSE, MAE, and $R^2$ values are averaged over three runs with different random seeds; lower RMSE/MAE and higher $R^2$ indicate better predictive performance.
For online learning methods, context length is also considered a tunable hyperparameter. 
All experiments are conducted on a single NVIDIA RTX A6000 GPU.

Due to confidentiality constraints in semiconductor manufacturing, the original sensor signals cannot be publicly released. 
Instead, we provide anonymized extracted feature vectors for each wafer, together with the corresponding target labels, as part of the code release for \ours.


\subsection{Comparison with Baseline Models}
\label{ssec:BenchmarkComparsion}



\mypar{Test Setting}
Table~\ref{tab:test-performance} shows that \ours achieves the best average rank among all compared methods, indicating consistent performance improvements across datasets and evaluation metrics. In particular, \ours improves the $R^2$ score by 2.8\% on D1 and 14.6\% on D2 compared to the best competitor, demonstrating its effectiveness under diverse distribution shifts.
The online learning baselines generally perform better than the batch methods, highlighting the importance of adaptive learning in non-stationary VM environments. However, their performance gains are unstable across datasets, while \ours consistently achieves strong performance due to the complementary design of the two adapters.

\mypar{Blind-Test Setting}
To reflect realistic deployment conditions with substantial future data, we conduct blind-test evaluation on the unseen split, where temporal distribution shift is most severe.
Table~\ref{tab:unseen-performance} shows that \ours achieves the best overall performance among the compared baselines.
AdaOTP, which integrates time-aware normalization with an adaptive ensemble of predictors, performs comparably to \ours on D1.
However, as it incrementally adds predictors during streaming updates, the model size and computational complexity grow over time. In contrast, \ours demonstrates strong robustness under varying distribution shifts while retaining a lightweight architecture.
Notably, consistent with Table~\ref{tab:test-performance}, the batch methods degrade substantially under severe distribution shifts, while online learning methods remain significantly stronger.

\begin{table}[t]
\caption{Ablation study on the test split, which shows the contribution of each component in \ours. 
The full \ours model generally achieves the best results.
}
\label{tab:ablation-study-test}
\centering
\large
\resizebox{\columnwidth}{!}{
\setlength{\tabcolsep}{6pt}
\renewcommand{\arraystretch}{1.05}
\begin{tabular}{l|ccc|ccc|c}
\toprule
\multirow{2}{*}{Method} 
& \multicolumn{3}{c|}{D1} 
& \multicolumn{3}{c|}{D2} 
& \multirow{2}{*}{\begin{tabular}{c}Avg \\ Rank\end{tabular}} \\

\cmidrule(lr){2-4}\cmidrule(lr){5-7}

& RMSE & MAE & $R^2$ & RMSE & MAE & $R^2$ & \\

\midrule

Global-MLP  & \underline{0.1649} & \textbf{0.1252} & \underline{0.1887} & \underline{0.8076} & 0.6485 & \underline{0.1891} & \underline{2.00}\\
Global-kNN  & 0.1651 & \underline{0.1254} & 0.1874 & 0.8321 & 0.6692 & 0.1392 & 3.50 \\
Adapter-I-RLS & 0.1695 & 0.1290 & 0.1430 & 0.8098 & \underline{0.6350} & 0.1847 & 3.33 \\
Adapter-II-None & 0.1718 & 0.1312 & 0.1196 & 1.0993 & 0.9012 & -0.5025 & 5.00 \\

\midrule
\textbf{\ours (ours)} & \textbf{0.1647} & 0.1270 & \textbf{0.1910} & \textbf{0.7658} & \textbf{0.5990} & \textbf{0.2709} & \textbf{1.17} \\

\bottomrule
\end{tabular}
}
\vspace{-5pt}
\end{table}







\begin{table}[t]
\caption{Ablation study on the blind-test split.
The full \ours model shows significant performance improvements over the four variants, especially on D1.}
\label{tab:ablation-study-unseen}
\centering
\large
\resizebox{\columnwidth}{!}{
\setlength{\tabcolsep}{6pt}
\renewcommand{\arraystretch}{1.05}
\begin{tabular}{l|ccc|ccc|c}
\toprule
\multirow{2}{*}{Method} 
& \multicolumn{3}{c|}{D1} 
& \multicolumn{3}{c|}{D2} 
& \multirow{2}{*}{\begin{tabular}{c}Avg \\ Rank\end{tabular}} \\

\cmidrule(lr){2-4}\cmidrule(lr){5-7}

& RMSE & MAE & $R^2$ & RMSE & MAE & $R^2$ & \\

\midrule

Global-MLP   & 0.7340 & 0.5118 & 0.4885 & 0.8149 & 0.6378 & 0.1758 & \underline{3.00} \\
Global-kNN   & \underline{0.7297} & \underline{0.5091} & \underline{0.4945} & 0.8225 & 0.6438 & 0.1602 & \underline{3.00} \\
Adapter-I-RLS & 0.7660 & 0.5223 & 0.4430 & \underline{0.8104} & \underline{0.6294} & \underline{0.1848} & \underline{3.00} \\
Adapter-II-None & 1.2331 & 0.8625 & -0.4435 & 1.1309 & 0.9171 & -0.5875 & 5.00 \\

\midrule
\textbf{\ours (ours)} & \textbf{0.2099} & \textbf{0.1540} & \textbf{0.9582} & \textbf{0.7962} & \textbf{0.6165} & \textbf{0.2132} & \textbf{1.00} \\

\bottomrule
\end{tabular}
}
\vspace{-10pt}
\end{table}








\subsection{Deeper Analysis on \ours}

\mypar{Case Study}
We present representative case studies in Fig.~\ref{fig:case-study} to qualitatively illustrate the adaptation behavior of \ours under temporal distribution shifts.
Fig.~\ref{fig:case-study-d1} is drawn from D1, where the target trajectory exhibits an abrupt regime change.
Linear and MLP, which are batch methods, remain close to their pre-shift predictions and fail to follow the new target level.
Rolling OLS reacts to the shift but exhibits large fluctuations and produces unstable predictions, while Online Linear adapts only partially and still leaves a noticeable residual gap. 
In contrast, \ours updates predictions after each streamed sample and progressively aligns its outputs with the shifted regime both rapidly and accurately.

Fig.~\ref{fig:case-study-d2} shows an example from D2, where the target trajectory changes more gradually due to covariate shift. 
\ours most effectively tracks the evolving dynamics of the target values compared to other baselines.
AdaOTP, which responds quickly to the target level shifts observed in D1, becomes less stable under the different shift patterns observed in D2.
This is because AdaOTP updates the predictive function itself during adaptation, which reduces its stability over time, while \ours effectively maintains the global predictor.

\mypar{Ablation Study}
To validate the design choices of \ours, we consider four variants: (1) replacing the global model with an MLP, (2) replacing the global model with a k-NN memory module, (3) replacing the NLMS update in Adapter~I with RLS, and (4) removing the offset module in Adapter~II. Tables~\ref{tab:ablation-study-test} and~\ref{tab:ablation-study-unseen} report the corresponding results on the test and blind-test splits, respectively.

On the test split, the full model achieves the best overall performance, confirming the effectiveness of the complete design.
Notably, removing Adapter II causes a substantial performance drop, indicating that the offset module plays a critical role in compensating for the environment-level bias.
On the blind-test split, the full model also achieves the best performance. In particular, on D1, which exhibits more dynamic shifts compared to D2, all variants perform substantially worse than the full model, with the best competing variant achieving only about half of the full model's performance.
Overall, these results indicate that each component of \ours contributes to its robust generalization.

\section{Conclusion}
\label{sec:conc}

In this work, we develop \ours, a practical online virtual metrology framework for maintaining reliable VM predictions in evolving semiconductor production environments.
The framework combines a stable global predictor built on interpretable feature construction that captures comprehensive wafer properties with lightweight streaming adapters that perform feature-level and environment-level corrections during inference.
By updating these lightweight adapters on a per-wafer basis, \ours adapts to feature-level and environment-level shifts without requiring full model retraining.
Comprehensive evaluations on two real-world semiconductor datasets with chronological test and blind-test splits show that \ours achieves the most reliable overall performance under realistic deployment conditions.
In particular, it outperforms representative batch and online baselines while maintaining strong robustness and balanced generalization. Ablation and 
case studies further confirm that the complementary adapter design plays a key role in enabling stable adaptation under diverse shift patterns.
These findings suggest that the proposed lightweight and interpretable online adaptation is a promising direction for robust VM deployment in evolving semiconductor manufacturing lines.

\ifCLASSOPTIONcompsoc
  \section*{Acknowledgments}
\else
  \section*{Acknowledgment}
\fi
This work was supported by Samsung Electronics Co., Ltd. (IO240426-09743-01) and conducted while the corresponding author was at KAIST.



%


\bibliographystyle{IEEEtran}
\bibliography{reference}

\end{document}